\documentclass[final]{cvpr}

\usepackage{times}
\usepackage{epsfig}
\usepackage{graphicx}
\usepackage{amsmath}
\usepackage{amssymb}

\usepackage[pagebackref=true,breaklinks=true,colorlinks,bookmarks=false]{hyperref}

\def\cvprPaperID{1} 
\def\confYear{CVPR 2021}

\begin{document}

\title{Rock Classification Based On Residual Networks}

\author{Sining Zhoubian \quad Yuyang Wang \quad Zhihuan Jiang\\
Tsinghua University\\
{\tt\small \{Zbsn21, yuyang-w21, jiang-zh21\}@mails.tsinghua.edu.cn}
}
\maketitle

\begin{abstract}
   Rock Classification is an essential geological problem since it provides important formation information.\ However, exploration on this problem using convolutional neural networks is not sufficient.\ To tackle this problem, we propose two approaches using residual neural networks. We first adopt data augmentation methods to enlarge our dataset.\ By modifying kernel sizes,  normalization methods and composition based on ResNet34, we achieve an accuracy of \textbf{70.1\%} on the test dataset, with an increase of \textbf{3.5\%} compared to regular Resnet34.\ Furthermore, using a similar backbone like BoTNet that incorporates multihead self attention, we additionally use internal residual connections in our model.\ This boosts the model's performance, achieving an accuracy of \textbf{73.7\%} on the test dataset.\ We also explore how the number of bottleneck transformer blocks may influence model performance. We discover that models with more than one bottleneck transformer block may not further improve performance.\ Finally, we believe that our approach can inspire future work related to this problem and our model design can facilitate the development of new residual model architectures.
\end{abstract}

\section{Introduction}

The rock classification task represents a fundamental assignment for geologists. However, relying solely on human visual identification not only demands a high level of expertise from human professionals but also consumes significant time and effort. Therefore, employing computer vision methods to accomplish rock classification becomes imperative. As far as researchers are aware, some studies have been conducted in this domain. However, the models constructed by these researchers exhibit less-than-optimal accuracy. This discrepancy prompts researchers to delve deeper into the project, exploring more robust models and techniques to enhance overall performance. 

The primary impediment to achieving higher accuracy is the insufficient data available. With a total of 53 rock varieties to classify, each having merely forty or fewer images for training, the dearth of data poses a significant challenge for training effective neural networks. To address this data limitation, the first adopted approach involves data augmentation. This technique aims to augment the available dataset artificially, mitigating the impact of limited training samples. 

Moreover, the selection of an appropriate base model is crucial. Traditional machine learning models like SVM or Random Forest may prove too simplistic for such a complex task, while some neural networks might be overly intricate, especially when faced with constraints in time and computing resources. Extensive literature research led to the choice of ResNet34 as the foundational model. ResNet34 addresses challenges involving the vanishing gradient problem, particularly critical in tasks with sparse data. Additionally, its straightforward architecture renders it both easily implementable and widely adopted in both research and practical applications, particularly in the realm of image classification.

Building upon ResNet34 as the fundamental model and employing data augmentation techniques, a series of research and experiments were conducted to enhance network performance. Noteworthy advancements were achieved through the incorporation and modification of ConvNeXt and BoTNet, both of which will be elucidated in detail in the subsequent sections.


\section{Related Work}

\subsection{Resnets}

In recent years, deep convolutional neural networks (CNNs) have emerged as powerful tools for various computer vision tasks. Among the plethora of architectures, ResNet-34 stands out as a cornerstone in the field of deep learning. Developed by Kaiming He et al., ResNet-34 is part of the ResNet (Residual Networks) family, which revolutionized the way deep networks are designed and trained.

The key innovation introduced by ResNet-34 lies in its incorporation of residual learning, a technique that facilitates the training of exceptionally deep networks. Traditional deep networks often encounter difficulties in training due to the vanishing or exploding gradient problem. ResNet-34 addresses this challenge by introducing residual blocks, allowing the network to learn residual functions with respect to the layer inputs. This architectural modification not only eases the training process but also enables the construction of deeper networks with improved performance.

ResNet-34 specifically consists of 34 layers, featuring a deep and sophisticated structure that captures intricate hierarchical representations of visual information. Leveraging skip connections, residual blocks, and global average pooling, ResNet-34 excels in extracting discriminative features, making it particularly well-suited for image classification, object detection, and semantic segmentation tasks.

As we delve into the intricate details of our research, the utilization of ResNet-34 serves as a fundamental component in our pursuit of advancing the state-of-the-art in computer vision. Its efficacy in handling complex visual data and its ability to facilitate the training of deep neural networks make ResNet-34 a pivotal architecture in contemporary deep learning research.

\subsection{ConvNeXt}

Liu Zhuang et al., drawing inspiration from the Swin Transformer, propose ConvNeXt \cite{liu2022convnet}. Through a series of comparative experiments, ConvNeXt demonstrates faster inference speed and higher accuracy compared to Swin Transformer under the same FLOPS.

The research, starting from ResNet-50 or ResNet-200, systematically incorporates ideas from Swin Transformer in five aspects: macro design, depth-wise separable convolution, inverted bottleneck layer, large convolutional kernel, and detailed design. Subsequently, training and evaluation are conducted on ImageNet-1K to derive the core structure of ConvNeXt. Additionally, the study includes detailed optimizations, such as replacing the ReLU function with GeLU, employing fewer activation functions and normalization layers, substituting Batch Normalization with Layer Normalization, and decomposing downsampling layers.

Compared to the complex Swin Transformer, ConvNeXt does not innovate in terms of structure or methods but instead utilizes existing structures and methods entirely. Its introduction further enhances the potential of traditional convolutional networks.

\subsection{BoTNet}

Aravind Srinivas et al.\ proposed a BoTNet \cite{srinivas2021bottleneck} based on multi-head self-attention mechanism (MHSA), replacing one of the convolutional layers in ResNet's last residual connection block with multi-head self-attention layer, hoping to improve the performance of pure convolutional networks. BoTNet, as a combination of pure convolutional network and Transformer architecture, combines the advantages of both.
The work explores the effect of the multihead self attention layer on network performance based on ResNet50 networks. It tries to replace the convolutional layer at different locations of the network with the multihead self attention layer and investigate its test accuracy on some data sets. After experiments, it found that replacing the last convolution layer of ResNet50 achieved the highest test accuracy. Comparing BotNet50 with ResNet50, the former achieves about 1\% to 5\% improvement over some image classification tasks. In this work, we adopt a similar model layout as backbone. However, we design a new residual connection method named internal residual connection and utilize this method in the bottleneck transformer block. We also explore how the number of bottleneck transformer blocks may influence the model's performance.

\section{Approach}

\subsection{Data Augmentation}
Upon acquiring the dataset for rock classification, a comprehensive strategy of data augmentation was employed to enhance the model's ability to generalize across various geological formations and conditions. 

The images were subjected to a series of transformations, including rotation, horizontal and vertical flipping, cropping, and scaling. These augmentations were strategically chosen to simulate the variability inherent in rock formations, accounting for differences in orientation and geological features. By introducing these variations during training, the model becomes more adept at recognizing and classifying rock specimens under diverse circumstances.

Furthermore, adjustments were made to image attributes such as brightness, contrast, saturation, and hue. This not only serves to simulate variations in lighting conditions but also ensures that the model is resilient to different environmental factors that may affect image quality. These fine-tuned augmentations contribute to a more robust and adaptable rock classification model.

\begin{figure}
    \centering
    \includegraphics[width=1\linewidth]{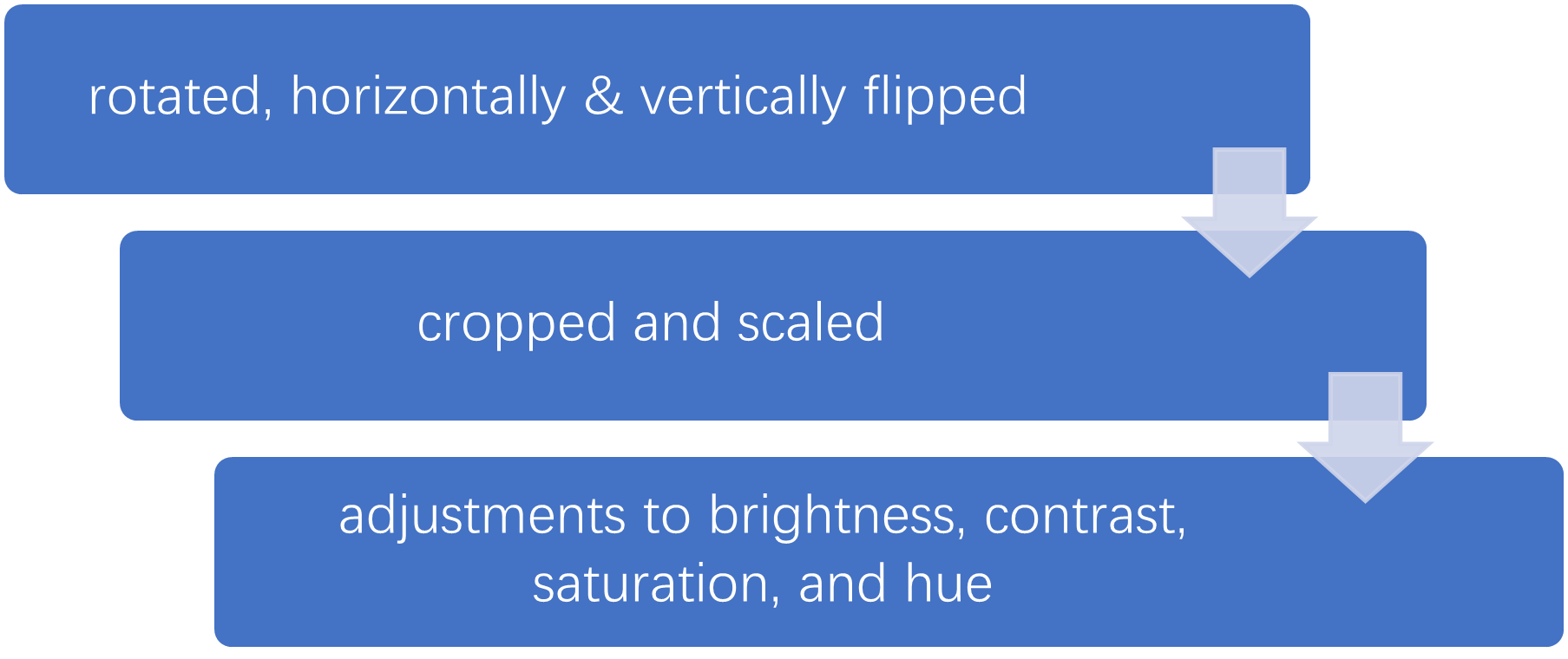}
    \caption{The data augmentation process}
    \label{fig:process}
\end{figure}

\begin{figure}
    \centering
    \includegraphics[width=1\linewidth]{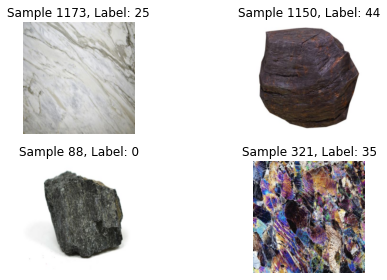}
    \caption{Some examples randomly extracted after data augmentation}
    \label{fig:examples}
\end{figure}

The process of the data augmentation and some examples extrated after data augmentation are shown in Figure \ref{fig:process} and Figure \ref{fig:examples}.


\subsection{Kernel Modification}
In this task, we drew inspiration from ConvNeXt's detailed optimization strategies. Specifically, we replaced the ReLU function with the GeLU function because the Gelu function is smoother and can to some extent mitigate the problem of gradient vanishing. We also employed fewer activation functions and normalization layers in the network, substituting Batch Normalization with Layer Normalization. These strategies were inspired by the principles of the Transformer. Additionally, we experimented with replacing the original BasicBlock in the ResNet-34 network with a BottleNeck structure, involving the addition of a 1x1 convolutional layer at the beginning of each residual block. This can help reduce computation and parameter count, effectively improving the model's performance and computational speed.

\subsection{Bottleneck Transformer}
By incorporating transformer's multihead self attention(MHSA) layer into pure CNNs, BoTNet outperforms ResNet in many image classification tasks, which confirms the feasibility to use MHSA in these tasks. In this work, we also test using MHSA in our model architecture. Specifically, we select the last two bottleneck blocks of ResNet34 as our main target to modify. We insert a MHSA layer into the two 3*3 kernel convolution layers with batch normalization and ReLU, forming a bottleneck transformer block.

\subsection{Internal Residual Connections}
\begin{figure}
    \centering
    \includegraphics[width=1\linewidth]{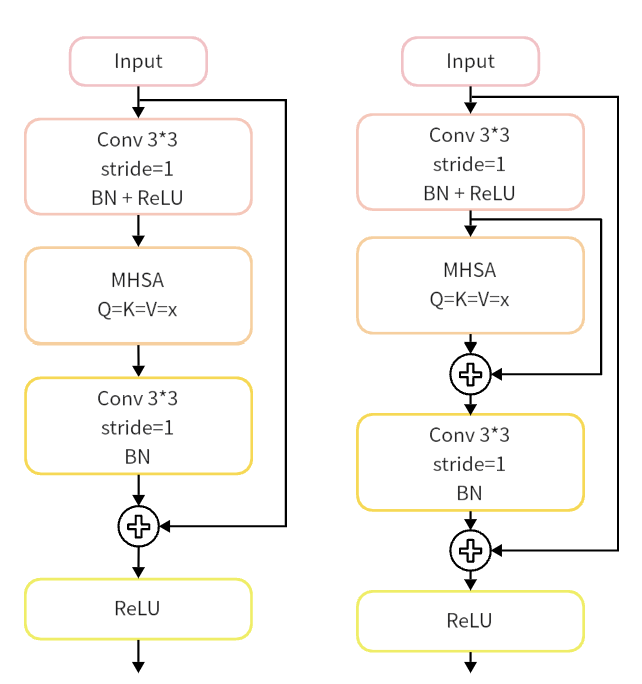}
    \caption{Layout of a bottleneck transformer block. \textbf{Left: }A normal bottleneck transformer block.  \textbf{Right: }A bottleneck transformer block with internal residual connection.}
    \label{fig:bottleneck transformer block}
\end{figure}

Although MHSA layers can enhance the model's performance in many circumstances, it often requires a large quantity of training data to fit. Besides, models combining convolution and attention may fit to datasets slower than common convnets. In some cases, these models may even not converge. To improve this situation, we design an internal residual connection(IRC) layout. While maintaining the convolution, MHSA layers and residual connection outside the whole bottleneck transformer block, we add a residual connection across the MHSA layer. This means the MHSA layer only needs to learn the second order difference of data, which simplifies its task and allows it to learn more subtle patterns in data. Layouts of common bottleneck transformer block and that with IRC are demonstrated in Figure \ref{fig:bottleneck transformer block}.

\section{Experiment}

\subsection{ResNet34 Implementation}
Despite the innate strengths of ResNet-34 in handling image classification challenges, its performance on our initial training set was surprisingly subpar, yielding an accuracy of a mere 30.64\%. Upon meticulous examination, it became apparent that the primary culprit behind this underwhelming outcome was the relatively small scale of our training dataset. Each rock type had, in all likelihood, only a handful of training samples—perhaps twenty or even fewer. Recognizing this limitation, we promptly implemented data augmentation as a strategic measure to address this deficiency.

\subsection{Data Augmentation Boosts Performance}
The culmination of these data augmentation techniques led to a significant improvement in the model's accuracy. The final accuracy achieved, after the incorporation of rotation, flipping, cropping, scaling, and attribute adjustments, reached an impressive 66.64\%, which is shown in Table \ref{tab:data augmentation}. This underscores the efficacy of data augmentation in addressing the challenges associated with rock classification, ultimately resulting in a more reliable and versatile model for geological image analysis.
\begin{table}
    \centering
    \begin{tabular}{|c|c|c|c|} \hline 
         Model &  Accuracy(\%)\\ \hline 
         Resnet-34 on initial dataset&  30.64\\ \hline 
         Resnet-34 with data augmentation&  66.64\\ \hline
    \end{tabular}
\vspace{\baselineskip}
    \caption{Results on test dataset after training 10 epochs using learning rate=1e-4. }
    \label{tab:data augmentation}
\end{table}

\subsection{Kernel Modification Improves Accuracy}
\begin{figure}
    \centering
    \includegraphics[width=0.5\linewidth]{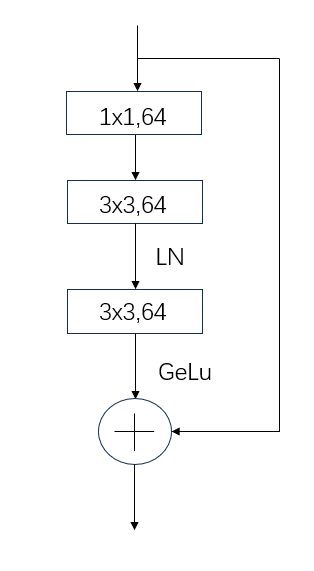}
    \caption{The structure of the modified kernel. }
    \label{fig:modified kernel}
\end{figure}

To validate the effectiveness of modifying internal details within residual blocks, we compared the performance of the original ResNet-34 network model with the modified network model. The modified residual block is illustrated in Figure \ref{fig:modified kernel}. Due to the significant improvement achieved by our data augmentation approach in performance, we applied this method to all models. The test results are presented in Table \ref{tab:modified kernel test results}. The results indicate that using the modified residual blocks can enhance the model's performance to a certain extent, achieving an accuracy of 70.1\%. This confirms the effectiveness of modifying the model based on ConvNeXt.

\begin{table}
    \centering
    \begin{tabular}{|c|c|c|c|} \hline 
         &  Accuracy(\%)\\ \hline 
         BasicBlock&  66.6\\ \hline 
         Replace Relu with Gelu&  66.6\\ \hline
         Fewer activation functions&  67.8\\ \hline
         Replace BN with LN&  68.3\\ \hline
         Add a 1x1 convolutional layer&  70.1\\ \hline
    \end{tabular}
\vspace{\baselineskip}
    \caption{Results on test dataset after training 10 epochs using learning rate=1e-4. }
    \label{tab:modified kernel test results}
\end{table}

\subsection{Bottleneck Transformer with Internal Residual Connection Achieves Top Accuracy }
\begin{figure}
    \centering
    \includegraphics[width=1\linewidth]{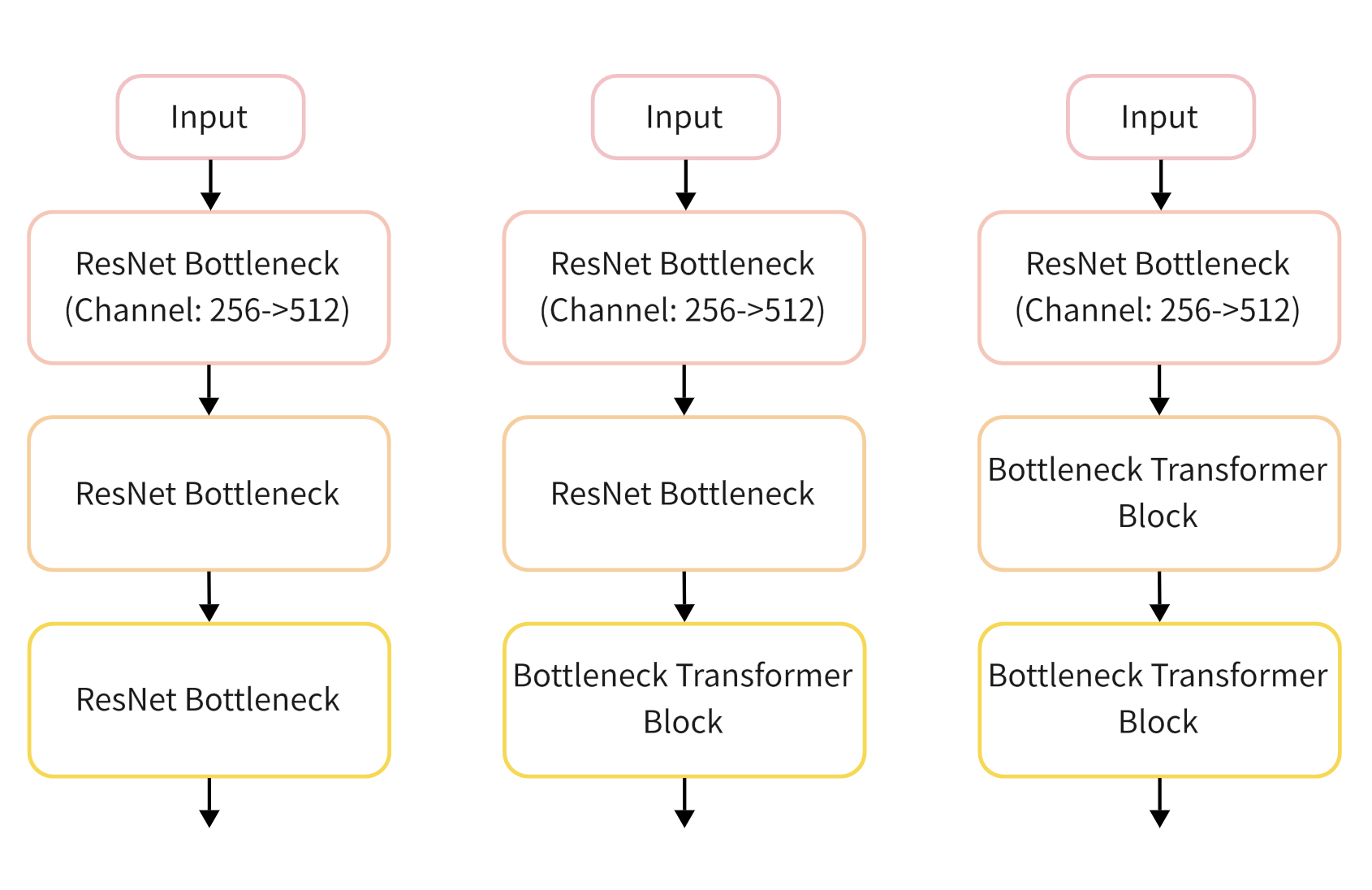}
    \caption{Layouts of the last convolution layer. \textbf{Left: }Original ResNet34 layout. \textbf{Middle: }One bottleneck transformer block. \textbf{Right: }Two bottleneck transformer blocks.}
    \label{fig:layer layout}
\end{figure}

To testify the effectiveness of bottleneck transformer and internal residual connection(IRC), we compare models with same ResNet34 backbone but different number of bottleneck transformer blocks (0, 1, 2) and different types of residual connection. All bottleneck transformer blocks are used in the last layer of ResNet34, replacing the original convolution bottleneck blocks. Layouts of the last layer are shown in Figure \ref{fig:layer layout}. Since our data augmentation approach shows significant improvement on performance, we adopt this approach for all models. Test results are shown in Table \ref{tab:BoT test results}. Results indicate that bottleneck transformer can have some positive effect when there is only one such block in the model layout, while more transformer blocks may impair performance. In contrast, IRC always improves models' performance, achieving the highest accuracy of 73.7\% with one bottleneck transformer block. This confirms the effectiveness of using IRC in networks that incorporate convolution and attention.

\begin{table}
    \centering
    \begin{tabular}{|c|c|c|c|} \hline 
         Accuracy(\%)&  0 BoT&  1 BoT& 2 BoT\\ \hline 
         Common&  66.6&  67.0& 65.9\\ \hline 
         Use IRC&  -&  73.7& 70.7\\ \hline
    \end{tabular}
\vspace{\baselineskip}
    \caption{Results on test dataset after training 10 epochs using learning rate=1e-4. }
    \label{tab:BoT test results}
\end{table}

\section{Conclusion}
Geological image identification and classification through deep learning is an underexplored field. In this work, we examine some approaches using residual networks as backbone. We discover that data augmentation can significantly improve accuracy on small datasets. It is verified that one bottleneck transformer block can enhance model's performance, while more of them may have a reverse effect. We also present internal residual connection, which is proved to have a positive effect on model fitting in all concerned cases. For future work, considering that related datasets are mostly insufficient, it is important to acquire some new data and expand the datasets. Besides, we haven't explored combining the two approaches together, namely kernel modification and bottleneck transformer with IRC. It may be a prospective approach for a wider range of tasks.

\section{Acknowledgement}
We sincerely appreciate Professor Xiaolin Hu and Assistant Professor Guo Chen for their guidance and assistance in selecting and conducting our research topic.

{\small
\bibliographystyle{ieee_fullname}
\bibliography{cvpr}
}

\end{document}